\documentclass[preprint,12pt]{elsarticle}

\usepackage{amssymb}
\usepackage{amsmath}

\usepackage{multirow}
\usepackage{booktabs}
\usepackage{CJKutf8}
\usepackage{amsmath,amssymb}
\usepackage{microtype}
\usepackage{stmaryrd}
\usepackage{cleveref}
\usepackage{mathtools}
\usepackage{verbatim}

\newcommand{\boxlr}{%
  \mathrel{\ooalign{%
    $\square$\cr
    \hidewidth\raisebox{0.38ex}{\scalebox{1.0}[0.65]{$|$}}\hidewidth\cr%
  }}%
}
\newcommand{\boxsurround}{%
  \mathrel{\ooalign{%
    $\square$\cr%
    \hidewidth\raisebox{0.35ex}{\scalebox{0.5}{$\square$}}\hidewidth\cr%
  }}%
}

\journal{Pattern Recognition}

\begin{document}

\begin{frontmatter}



\title{Entropy-Aware Structural Alignment for Zero-Shot Handwritten Chinese Character Recognition}


\author[1,2]{Qiuming Luo}
\ead{lqm@szpu.edu.cn}

\author[1]{Tao Zeng}
\ead{zengtao2023@email.szu.edu.cn}

\author[3]{Feng Li}
\ead{lifeng9472@szpu.edu.cn}

\author[3]{Heming Liu}
\ead{liuheming@szpu.edu.cn}

\author[1]{Rui Mao}
\ead{mao@szpu.edu.cn}

\author[3,4]{Chang Kong\corref{cor1}}
\ead{kongchang@szpu.edu.cn}

\cortext[cor1]{Corresponding author.}

\affiliation[1]{organization={College of Computer Science and Software Engineering, Shenzhen University},
            addressline={Nanshan}, 
            city={Shenzhen},
            postcode={518060}, 
            state={Guangdong},
            country={China}}

\affiliation[2]{organization={Shenzhen Key Laboratory of Embedded System Design},
            addressline={Nanshan}, 
            city={Shenzhen},
            postcode={518060}, 
            state={Guangdong},
            country={China}}
            
\affiliation[3]{organization={Undergraduate School of Artificial Intelligence, Shenzhen Polytechnic University},
                addressline={Nanshan},
                city={Shenzhen},
                postcode={518055}, 
                city={Shenzhen},
                state={Guangdong}, 
                country={China}}

\affiliation[4]{organization={Institute of Applied Artificial Intelligence of the Guangdong-Hong Kong Macao Greater Bay Area},
                addressline={Nanshan}, 
                city={Shenzhen},
                postcode={518055}, 
                state={Guangdong}, 
                country={China}}

\begin{abstract}
Zero-shot Handwritten Chinese Character Recognition (HCCR) aims to recognize unseen characters by leveraging radical-based semantic compositions. However, existing approaches often treat characters as flat radical sequences, neglecting the hierarchical topology and the uneven information density of different components. To address these limitations, we propose an Entropy-Aware Structural Alignment Network that bridges the visual-semantic gap through information-theoretic modeling. First, we introduce an Information Entropy Prior to dynamically modulate positional embeddings via multiplicative interaction, acting as a saliency detector that prioritizes discriminative roots over ubiquitous components. Second, we construct a Dual-View Radical Tree to extract multi-granularity structural features, which are integrated via an adaptive Sigmoid-based gating network to encode both global layout and local spatial roles. Finally, a Top-K Semantic Feature Fusion mechanism is devised to augment the decoding process by utilizing the centroid of semantic neighbors, effectively rectifying visual ambiguities through feature-level consensus. Extensive experiments demonstrate that our method establishes new state-of-the-art performance, achieving an accuracy of 55.04\% on the ICDAR 2013 dataset ($m=1500$), significantly outperforming existing CLIP-based baselines in the challenging zero-shot setting. Furthermore, the framework exhibits exceptional data efficiency, demonstrating rapid adaptability with minimal support samples, achieving 92.41\% accuracy with only one support sample per class.
\end{abstract}




\end{frontmatter}



\section{Introduction}\label{sec:intro}
Handwritten Chinese Character Recognition (HCCR) has long been a fundamental challenge in the field of pattern recognition and document analysis. Unlike alphabetic scripts, Chinese characters are logographic, characterized by highly complex topological structures and a vast vocabulary size—the GB18030-2022 standard alone encodes over 80,000 characters \cite{zhang2017}. While deep learning-based methods have achieved remarkable success on standard benchmarks with fixed characters \cite{yin2013,xiao2017}, they often struggle with the \textit{open-set recognition problem}, where the model encounters characters during testing that were never seen during training. This limitation has spurred significant research interest in Zero-Shot Learning (ZSL) for HCCR, which aims to recognize unseen characters by transferring knowledge from seen classes via intermediate semantic representations \cite{wang2018,chen2021a}.

To bridge the gap between visual features and semantic concepts in ZSL, decomposing characters into radicals or Ideographic Description Sequences (IDS) has become the prevailing strategy \cite{zhang2020,chen2021a}. Radicals serve as the "atomic" units of Chinese characters, allowing unseen characters to be recognized by identifying their constituent parts and structural layouts. Recent advancements have further leveraged large-scale Vision-Language Models (VLMs), such as CLIP \cite{radford2021}, to learn robust multimodal representations. For instance, CCR-CLIP \cite{yu2023a} aligns glyphs with radical sequences, and more recently, Cai et al. \cite{cai2024b} proposed a cross-modal framework that aligns local discrete tokens with global character features. In our preliminary work, we demonstrated that fine-tuning CLIP for radical encoding can effectively align visual glyphs with their textual descriptions, achieving promising results in zero-shot scenarios \cite{luo2025}.

However, despite these advancements, existing radical-based approaches, including our previous work, exhibit two critical limitations that hinder further performance gains:

\begin{CJK*}{UTF8}{gbsn}
First, \textit{information inequality among radicals} is largely overlooked. Most existing methods \cite{cao2020,zhang2020} treat every radical in a sequence as an equally informative token. From an information-theoretic perspective \cite{shannon1948}, this assumption is flawed. High-frequency radicals (e.g., `口' (mouth) or `日' (sun)) appear in thousands of characters and thus carry low entropy, offering limited discriminative power for fine-grained classification. Conversely, rare radicals often serve as unique identifiers for specific characters. Indiscriminately aggregating these features dilutes the contribution of key discriminative components, leading to confusion among visually similar characters (e.g., ``货'' (goods) vs. ``贷'' (loan)).
\end{CJK*}

Second, the \textit{alignment between visual features and semantic embeddings} remains coarse. Previous methods typically rely on simple mechanisms, such as cosine similarity or linear concatenation, to fuse visual features with radical embeddings \cite{wang2019,yu2023a}. This shallow interaction fails to capture the complex, non-linear dependencies between the distorted strokes of handwriting and the rigid definitions of radical structures. Furthermore, the structural information of Chinese characters is inherently hierarchical—comprising both global layouts (e.g., left-right, surround) and local compositions \cite{li2024o}. Existing flat or simple tree-based embeddings fail to explicitly model these dual-view structural dependencies, resulting in a loss of spatial context.

To address these challenges, we propose a novel framework titled \textbf{Entropy-Aware Structural Alignment (EASA)} network. Specifically, we first introduce an Entropy-Aware Position Embedding (EAPE) mechanism. By quantifying the information entropy of each radical based on its statistical frequency, we dynamically modulate the positional embeddings via multiplicative interaction, forcing the model to attend more to high-entropy, discriminative components while suppressing redundant high-frequency signals. Secondly, we upgrade the structural representation by constructing Dual-View Radical Trees, which explicitly aggregate multi-granularity structural features from two distinct perspectives: a parent-centric view capturing global layout dependencies and a child-centric view preserving local compositional details. Furthermore, to achieve precise cross-modal alignment, we propose a Radical Semantic Matching Module. Unlike simple similarity metrics employed in prior work \cite{cao2020,yu2023a}, this module employs an adaptive Sigmoid-based GateFusion network to integrate heterogeneous structural features and utilizes a Cross-Modal Attention mechanism \cite{vaswani2017,lee2018} to perform deep reasoning between visual and semantic representations. Finally, we introduce a Top-K Semantic Feature Fusion strategy, which augments the decoder's query with the centroid of semantic neighbors to improve robustness against handwriting ambiguity.

The main contributions of this work are summarized as follows:
\begin{itemize}
    \item \textbf{Entropy-Guided Representation}: We propose a novel entropy-aware modulation scheme that utilizes multiplicative interaction to theoretically quantify and leverage the discriminative value of radicals, effectively resolving the information imbalance problem in radical sequence modeling.
    \item \textbf{Dual-View Structural Embedding}: We develop a hierarchical tree embedding method that extracts five distinct multi-granularity features to simultaneously model global parent-child dependencies and local structural contexts, providing a richer structural prior than simple sequence or single-view tree models.
    \item \textbf{Multi-Stage Semantic Matching}: We design a comprehensive matching module integrating Sigmoid-based GateFusion and Top-K Feature Fusion to enable explicit and deep alignment between handwritten visual features and semantic priors.
    \item \textbf{Extensive Evaluation}: Beyond the zero-shot settings, we conduct extensive experiments on few-shot learning, full-set recognition, and detailed visualizations (e.g., entropy heatmaps and matching trajectories). Experimental results on CASIA-HWDB and ICDAR2013 datasets demonstrate that our method achieves state-of-the-art performance.
\end{itemize}

\section{Related Work}\label{sec:related}
This section reviews the literature relevant to our proposed framework, focusing on three key aspects: zero-shot Handwritten Chinese Character Recognition, radical-based structural representation, and cross-modal semantic alignment.

\subsection{Zero-shot HCCR via Decomposition}
Traditional HCCR methods predominantly rely on deep convolutional neural networks (CNNs) trained on large-scale datasets with fixed characters \cite{shi2017,xiu2019}. While these fully supervised models achieve high accuracy on closed sets, they fail to generalize to the open-set nature of Chinese characters, where thousands of rare characters and newly coined glyphs exist outside standard training sets \cite{zhang2017}.

To address this, zero-shot learning has been introduced to recognize unseen characters by exploiting the compositional nature of Chinese characters. The prevailing strategy involves decomposing characters into smaller semantic units—radicals or strokes—and learning a mapping between visual features and these sub-character sequences \cite{chen2021a}. Early works, such as the DenseRAN \cite{wang2018}, treated radical sequences as text strings, utilizing Recurrent Neural Networks (RNNs) to generate captions for character images. Subsequent approaches, like HDE \cite{cao2020} and RAN \cite{zhang2020}, incorporated two-dimensional spatial attention to better capture the spatial arrangement of radicals. More recently, Diao et al. \cite{diao2023} proposed RZCR, which introduces a knowledge graph reasoner to explicitly model the logical relations between radicals for zero-shot inference.

\begin{CJK*}{UTF8}{gbsn}
However, most existing ZSL frameworks treat the decomposition as a static look-up process. They often rely on one-hot or few-hot encodings of radicals, which result in sparse high-dimensional spaces and fail to capture the semantic similarities between visually akin radicals (e.g., `讠' (speech) and `氵' (water)). Although some works have explored continuous representations, such as stroke-level embeddings in CW2Vec \cite{cao2018} or radical aggregation in RAN \cite{wang2019}, they typically lack a mechanism to explicitly model the discriminative importance of different radicals, treating common ubiquitous components and rare identifiers with equal weight.
\end{CJK*}

\subsection{Radical Representation and Structure Modeling}
The efficacy of ZSL depends heavily on how the internal structure of a character is modeled. While early sequence-based methods treated a character as a 1D string of radicals to leverage standard RNNs \cite{wang2018,zhang2020}, they inevitably lose critical topological information, such as the spatial hierarchy of surround, above-below, and left-right structures.

To preserve topological fidelity, tree-structured representations have emerged as a superior alternative. Methods like HDE \cite{cao2020} and SideNet \cite{li2024o} employ radical trees to model the hierarchical recursive composition of characters. Hong et al. \cite{hong2025} proposed FT-CLIP, which further validates the efficacy of formation trees by integrating them with a masked image modeling paradigm. In our preliminary work \cite{luo2025}, we advanced this direction by fine-tuning the CLIP model to learn visual-aligned embeddings for radical trees, effectively bridging the gap between glyphs and their IDS.

\begin{CJK*}{UTF8}{gbsn}
Despite these structural advancements, a critical theoretical gap remains: the neglect of \textit{information entropy}. Current structural encoders, typically based on Tree-LSTMs \cite{tai2015} or Graph Convolutional Networks (GCNs) \cite{kipf2017}, generally assume that all nodes in the character tree contribute equally to the final representation. From an information-theoretic standpoint \cite{shannon1948}, this is suboptimal. High-frequency radicals (e.g., structural particles like `口') carry low information entropy and are less informative for distinguishing fine-grained characters, whereas low-frequency radicals often serve as key discriminators.
\end{CJK*}

\subsection{Cross-Modal Alignment and Matching Mechanisms}
Zero-shot recognition is fundamentally a cross-modal matching problem, aligning visual queries with semantic prototypes. The standard paradigm involves projecting visual features and semantic embeddings into a shared latent space and minimizing a distance metric, typically Euclidean distance \cite{frome2013} or Cosine similarity \cite{cao2020,yu2023a}.

While computationally efficient, these shallow metric learning approaches are often insufficient for HCCR. Handwriting introduces significant elastic deformations, stroke aliasing, and topological variability, leading to a severe domain shift between the noisy visual input and the rigid, canonical semantic definitions. As noted in recent studies \cite{li2025e}, simple scalar similarity metrics cannot capture the complex, non-linear alignments required to robustly match a distorted handwritten glyph to its structural prototype.

Recent advances in Vision-Language Pre-training (VLP) have demonstrated the power of Cross-Attention mechanisms for fine-grained image-text alignment. Models like CLIP \cite{radford2021} use contrastive learning for global alignment, while subsequent works like SCAN \cite{lee2018} and ALBEF \cite{li2021} employ deep cross-attention to allow the model to dynamically query relevant semantic components based on visual cues. Cai et al. \cite{cai2024b} introduced learnable tokens to bridge the gap between local radical features and global character embeddings. However, applying such deep interaction mechanisms to HCCR is non-trivial due to the hierarchical nature of Chinese characters. Simply concatenating visual and semantic features often leads to modality dominance, where the stronger modality (usually the pre-trained visual backbone) overshadows the weaker one \cite{wei2025b}.

\begin{figure}[t!]
    \centering
    \includegraphics[width=0.95\textwidth]{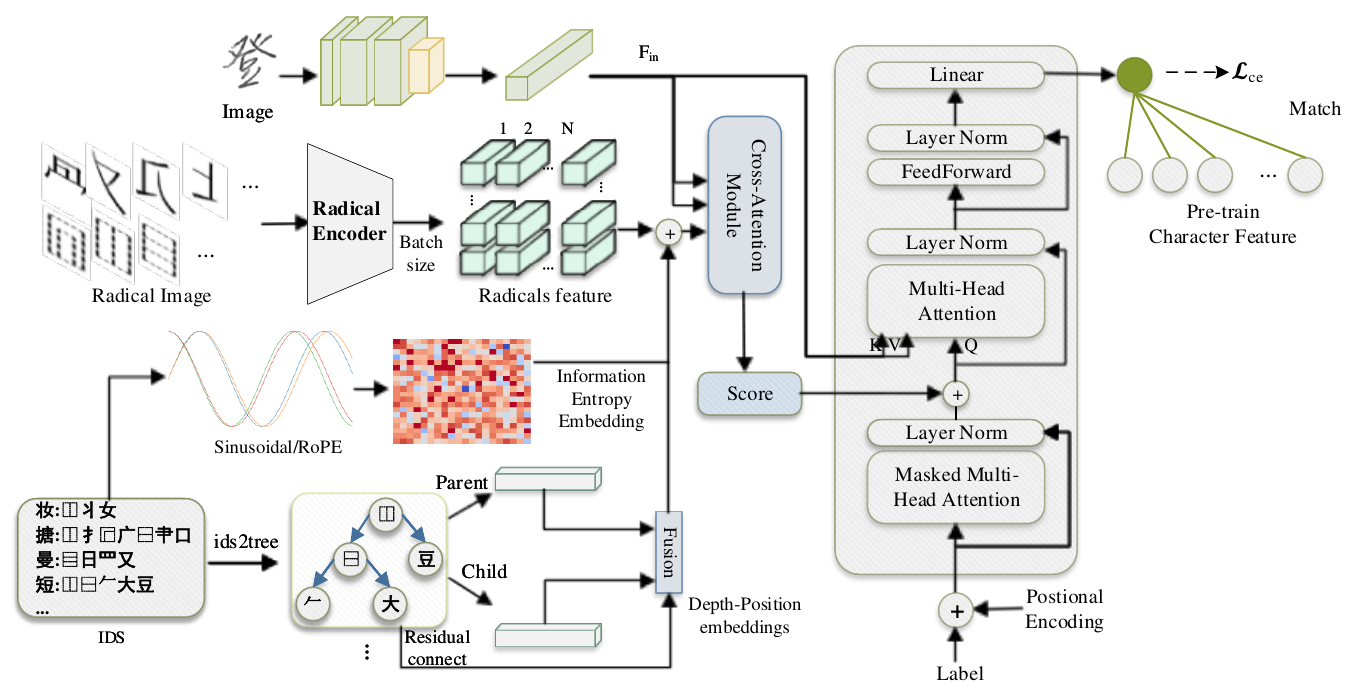}
    \vspace{-10pt}
    \caption{The overall architecture of the proposed Entropy-Aware Structural Alignment Network. The framework consists of three input branches and a central matching mechanism: 
    (1) The \textbf{Visual Branch} (top-left) employs a ResNet-based backbone to extract feature maps from handwritten character images. 
    (2) The \textbf{Radical Image Branch} (middle-left) utilizes our Multimodal Radical Encoder to extract visual-aligned radical features. 
    (3) The \textbf{IDS Branch} (bottom-left) constructs semantic representations using \textbf{Entropy-Aware Multiplicative Modulation} and \textbf{Dual-View Radical Tree Embeddings}. 
    The core component (highlighted in blue) is the \textbf{Radical Semantic Matching Module}, which aligns visual features with \textbf{five distinct structural representations} via an \textbf{Adaptive Sigmoid-GateFusion}. Furthermore, a \textbf{Top-K Semantic Feature Fusion} strategy is employed to \textbf{augment the decoder's query} with robust semantic priors. Finally, the \textbf{Transformer Decoder} (right) processes the enhanced features to generate the final recognition result.}
    \label{fig:framework}
    \vspace{-10pt}
\end{figure}

\section{Methodology}\label{sec:method}
In this section, we present the proposed Entropy-Aware Structural Alignment Network for zero-shot handwritten Chinese character recognition. As illustrated in \Cref{fig:framework}, the framework comprises two parallel distinct pathways: a visual stream and a semantic stream, which interact through a deep matching mechanism.

\subsection{Preliminaries}\label{subsec:prelim}
Before detailing our core contributions, we first formalize the zero-shot recognition problem and introduce the foundational components of our framework, including the specialized data augmentation strategy and the fine-tuned radical encoder adopted from our preliminary work \cite{luo2025}.

\subsubsection{Problem Formulation}
Let $\mathcal{D}_s = \{(x_i, y_i)\}_{i=1}^{N_s}$ denote the source training set (seen classes), where $x_i$ represents a handwritten character image and $y_i \in \mathcal{Y}_s$ is the corresponding label. The target testing set is denoted as $\mathcal{D}_u = \{(x_j, y_j)\}_{j=1}^{N_u}$ (unseen classes), where $y_j \in \mathcal{Y}_u$. In the zero-shot setting, the seen and unseen label sets are disjoint, i.e., $\mathcal{Y}_s \cap \mathcal{Y}_u = \emptyset$.

Our goal is to learn a mapping function $f: \mathcal{X} \rightarrow \mathcal{Y}$ capable of recognizing characters in $\mathcal{Y}_u$ by leveraging auxiliary semantic information. For each character class $c$, we utilize its radical sequence $R_c = [r_1, r_2, ..., r_L]$ and structural description as the semantic prototype. The model must generalize to $\mathcal{Y}_u$ by aligning the visual features of $x$ with the semantic representation of $R_c$.

\subsubsection{Multi-grid 2D Elastic Deformation}
Handwritten Chinese characters exhibit complex internal structures with non-rigid variations, such as stroke squeezing and local warping, which simple affine transformations fail to model \cite{simard2003}. Traditional augmentation methods often rely on 1D strip-based deformations \cite{luo2020}, which are insufficient for capturing the intricate stroke interplay of Chinese radicals.

\begin{figure}[t!]
    \centering
    \includegraphics[width=0.75\textwidth]{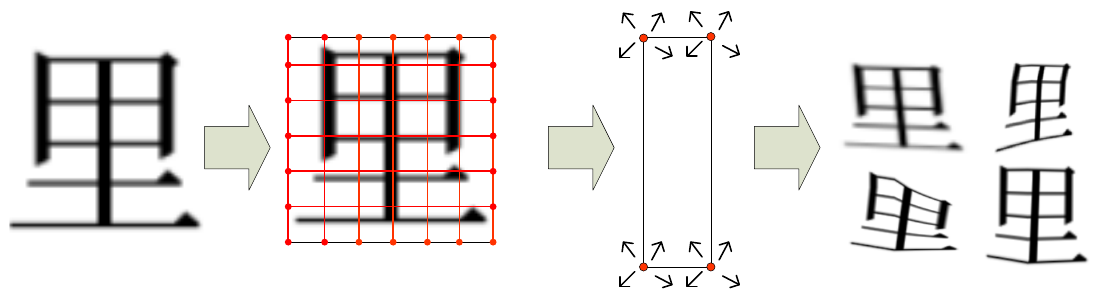}
    \vspace{-10pt}
    \caption{Illustration of the \textbf{Multi-grid 2D Elastic Deformation}. A dense 2D elastic mesh is constructed over the radical image, where control points $p_{m,n}$ (visualized as red dots) are independently perturbed.}
    \label{fig:augmentation}
    \vspace{-10pt}
\end{figure}

To address this, we propose a Multi-grid 2D Elastic Deformation strategy, as illustrated in \Cref{fig:augmentation}. Unlike coarse 1D methods, our approach constructs a fine-grained elastic mesh over the radical image, extending the deformation freedom to both horizontal and vertical dimensions. Specifically, we define a regular grid $G = \{p_{m,n}\}$ over the input image $I$, where $p_{m,n}$ denotes the control points. We generate a random displacement field $\Delta G$ by sampling offsets from a Gaussian distribution $\mathcal{N}(0, \sigma^2)$ for each control point. The deformed image $I'$ is then generated by mapping the pixel locations using bicubic interpolation based on the displaced grid $G + \Delta G$. This process generates realistic "wobbly" strokes and structural topological variations, forcing the network to learn invariant features robust to severe handwriting distortions.

\subsubsection{Fine-tuned CLIP for Radical Representation}\label{subsubsec:rad-clip}
To bridge the modality gap between visual glyphs and textual radicals, we adopt the fine-tuning strategy from our previous work \cite{luo2025}. Instead of initializing radical embeddings randomly or using standard word vectors (e.g., Word2Vec), we leverage the CLIP model, which has learned rich visual-semantic alignments.

For each unique radical $r$ in the vocabulary, we construct a prompt template ``A photo of [radical]'' and feed it into the CLIP text encoder to obtain an initial continuous embedding $e_r^{\text{clip}} \in \mathbb{R}^d$. To adapt these generic embeddings to the specific domain of Chinese calligraphy and handwriting, we employ a prompt tuning mechanism where learnable context vectors are optimized alongside the network. This results in a comprehensive radical embedding dictionary $\mathbf{E}_{\text{rad}}$, which serves as the fundamental input for our subsequent entropy-aware and structural modules.

\subsection{Entropy-Aware Position Embedding}\label{subsec:entropy-aware}
Standard positional encodings (PE) in Transformer-based architectures, ranging from absolute embeddings \cite{vaswani2017} and relative position representations \cite{shaw2018} to recent rotary mechanisms like RoFormer \cite{su2024}, serve a singular purpose: to inject sequence order information into permutation-invariant self-attention mechanisms. However, in the context of Chinese character modeling, treating position information purely as a geometric index—whether absolute or relative—is suboptimal. This is because the informational value of radicals at different positions is highly non-uniform.

\begin{CJK*}{UTF8}{gbsn}
According to Information Theory, the information content of a token is inversely proportional to its probability of occurrence. In the decomposition of Chinese characters, radicals follow a heavy-tailed distribution (Zipf's law). High-frequency radicals (e.g., `口' (mouth) or `亻' (human)) appear in thousands of characters, serving as structural particles with low discriminative power (low entropy). Conversely, low-frequency radicals often act as unique identifiers for specific characters (high entropy). Existing methods that utilize standard PE or uniform attention inherently neglect this information inequality, allowing high-frequency redundant signals to dominate the feature space.
\end{CJK*}

To address this, we propose an Entropy-Aware Position Embedding mechanism. This module explicitly quantifies the information density of each radical and uses a multiplicative interaction mechanism to dynamically modulate the embedding space.

\subsubsection{Information Entropy Quantification}
Formally, let $\mathcal{V} = \{r_1, r_2, \dots, r_M\}$ be the vocabulary of all unique radicals in the dataset. We first compute the global occurrence frequency $N(r_k)$ for each radical $r_k \in \mathcal{V}$ across the entire training corpus. Specifically, the probability of the $k$-th radical is calculated as $P(r_k) = N(r_k) / \sum_{j=1}^{M} N(r_j)$, where $N(r_k)$ denotes the frequency of radical $r_k$ in the training corpus and $M$ is the vocabulary size. Distinct from standard definitions that typically employ base-2 logarithms (bits), we adopt the natural logarithm to measure uncertainty (nats), aligning better with the optimization dynamics of the continuous representation space. The Information Entropy is defined as $H(r_k) = -\ln(P(r_k))$, which ensures that ubiquitous radicals yield a low entropy value while rare, discriminative radicals yield a high value.

\begin{figure}[t!]
    \centering
    \includegraphics[width=0.5\textwidth]{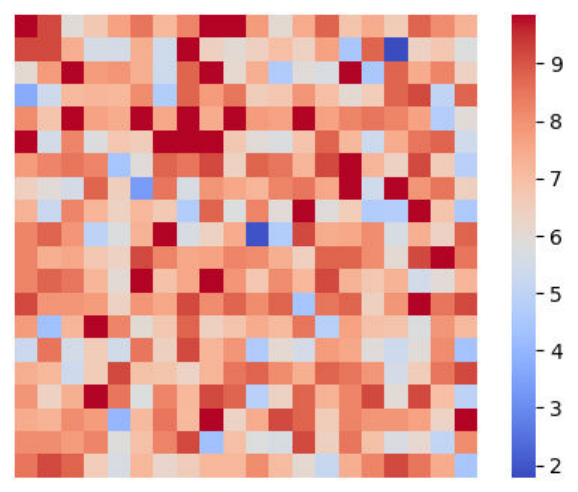}
    \vspace{-10pt}
    \caption{Visualization of Information Entropy Statistics for over 400 radicals. The heatmap spectrum ranges from deep blue (low entropy) to deep red (high entropy), illustrating the non-uniform distribution of information density across the vocabulary.}
    \label{fig:entropy_stat}
    \vspace{-10pt}
\end{figure}

To empirically validate the necessity of this modeling, we visualized the entropy distribution of over 400 radicals in our dataset, as shown in \Cref{fig:entropy_stat}. The visualization reveals a distinct ``long-tail'' characteristic: deep red blocks (entropy $\approx 9$) represent rare radicals with high information content, while the few deep blue blocks (entropy $\approx 2$) correspond to ubiquitous components with low discriminative power. The statistical evidence that radical information is unevenly distributed justifies our strategy to dynamically re-weight components based on their information density.

\subsubsection{Embedding Modulation and Aggregation}
We integrate this entropy prior directly into the positional embedding process via a multiplicative gating mechanism, rather than the standard additive approach.

Let $\mathbf{e}_i \in \mathbb{R}^d$ denote the semantic embedding of the $i$-th radical, \textbf{retrieved from the pre-computed dictionary $\mathbf{E}_{rad}$} (defined in \Cref{subsubsec:rad-clip}), and $\mathbf{p}_i \in \mathbb{R}^d$ be the standard learnable positional embedding. The entropy-aware representation $\mathbf{v}_i$ is computed as:
\begin{equation}
\mathbf{v}_i = \mathbf{e}_i \odot (H(r_i) \cdot \mathbf{p}_i)
\label{eq3:vi_entropy}
\end{equation}
where $\odot$ denotes the element-wise product (Hadamard product), and $H(r_i)$ is the scalar entropy value. In this formulation, the positional signal is first scaled by the radical's entropy intensity ($H(r_i)$). High-entropy radicals result in stronger positional signals, while low-entropy radicals have their spatial influence dampened. This scaled position vector is then fused with the semantic embedding $\mathbf{e}_i$ via element-wise multiplication, effectively creating an attention mask that highlights discriminative structural components.

Finally, to capture the global structural signature, we aggregate the sequence into a global representation vector $\mathbf{v}_{ent} = (1/L) \sum_{i=1}^{L} \mathbf{v}_i$. This mean-pooling operation ensures robustness to sequence length variations while preserving the centroid of the entropy-weighted semantic space, aligning the network's focus with human cognitive intuition.

\subsection{Dual-View Structural Representation}\label{subsec:tree}
\begin{CJK*}{UTF8}{gbsn}
While the entropy-aware embedding described in \Cref{subsec:entropy-aware} effectively highlights discriminative components, it operates primarily on a sequential level. However, Chinese characters are inherently hierarchical 2D structures. For instance, the character ``森'' (forest) is not merely a linear sequence of three ``木'' (wood) radicals; it is structurally composed of one ``木'' on top and two ``木''s below. Neglecting this topological layout leads to ambiguity among characters that share identical radical components but differ in spatial arrangement (e.g., ``呆'' vs. ``杏'').
\end{CJK*}

To rigorously capture this structural topology, we propose a Dual-View Radical Tree (DVRT) mechanism. We first parse the IDS into a binary syntax tree, where leaf nodes represent radicals and internal nodes represent structural operators (e.g., $\text{Left-Right } \boxlr$, $\text{Above-Below } \boxminus$, $\text{Surround } \boxsurround$). Based on this topology, we compute embeddings from two distinct perspectives: a Parent-Centric Global View and a Child-Centric Local View, ultimately extracting five distinct structural feature vectors to comprehensively describe a character's topological identity. From a graph theory perspective, this dual-view mechanism can be theoretically viewed as decomposing the character's semantic graph into a Directed Acyclic Graph (DAG) of dependencies (Parent-View) and a set of positional attributes (Child-View).

\begin{figure}[t!]
    \centering
    \includegraphics[width=0.8\linewidth]{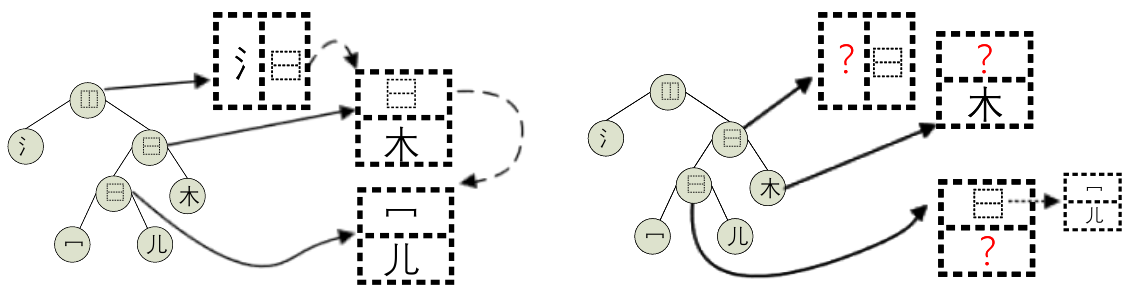} 
    \vspace{-15pt}
    \begin{CJK*}{UTF8}{gbsn}
    \caption{Illustration of the \textbf{Dual-View Radical Tree Modeling} using the character ``深'' (deep). 
    \textbf{(Left) Parent-centric Global View:} Depicts bottom-up aggregation, where the root `$\protect\boxlr$' aggregates semantic features from its left child `氵' and right child (composed of `穴' and `木'), capturing the global composition. 
    \textbf{(Right) Child-centric Local View:} Depicts local topological roles, where a node (e.g., `木') explicitly attends to its direct parent (`$\protect\boxminus$') to identify its specific structural position (e.g., bottom child).}
    \label{fig:tree_process}
    \end{CJK*} 
    \vspace{-10pt}
\end{figure}

\subsubsection{Tree Construction and Depth-Position Embedding}
\begin{CJK*}{UTF8}{gbsn}
To capture the intrinsic hierarchical topology, we adopt the tree parsing protocol from our preliminary work \cite{luo2025}. As visually demonstrated in \Cref{fig:tree_process} using the character ``\begin{CJK*}{UTF8}{gbsn}深\end{CJK*}'' as an example, the radical sequence is constructed into a binary tree $\mathcal{T}$, where each node $n_i$ is assigned a depth $l_i$ (root at 0) and a branch position indicator $pos_i \in \{0, 1, 2\}$ (0 for root, 1 for left child, 2 for right child).
\end{CJK*}

To encode these spatial coordinates, we generate a Depth-Position ($DP$) embedding for each node. Unlike standard sinusoidal encodings, our $DP$ embedding simultaneously encodes the node's depth and its branch position:
\begin{equation}
\mathbf{DP}_i = 
\begin{cases} 
\sin\left(\frac{2 \cdot d \cdot \pi}{D}\right), & \text{if } pos_i = 0 \\
\sin\left(\frac{(4l_i - 2) \cdot d \cdot \pi}{D}\right), & \text{if } pos_i = 1 \\
\sin\left(\frac{4l_i \cdot d \cdot \pi}{D}\right), & \text{if } pos_i = 2 
\end{cases}
\label{eq5:dp_embedding}
\end{equation}
where $d$ is the dimension index and $D=512$ is the embedding dimension. This ensures that nodes at different depths and branches are mapped to orthogonal regions in the embedding space.

\subsubsection{Parent-Centric Global View}
The global view aims to encapsulate the holistic structure of the character. We employ a bottom-up recursive aggregation strategy, where the representation of a parent node accumulates positional context from its ancestors.

For a node $n_i$ at depth $l_i$, its parent-centric embedding $\mathbf{DP}_{parent, i}$ is computed as:
\begin{equation}
\mathbf{DP}_{parent, i} = \mathbf{DP}_i + \frac{1}{l_i} \sum_{k=1}^{l_i} \sin\left(\frac{4 \cdot k \cdot d \cdot \pi}{D}\right)
\label{eq6:dp_parent}
\end{equation}
This summation captures the global layout path from the root to the current node, enabling the representation to reflect the node's absolute position within the character's overall framework.

\subsubsection{Child-Centric Local View}
While the global view aggregates information, it may obscure the individual role of specific radicals within the local substructure. To preserve these details, we introduce a child-centric view. 

This view explicitly encodes the structural role of each node relative to its immediate siblings or children. The child-centric embedding $\mathbf{DP}_{child, i}$ is computed by aggregating the $DP$ embeddings of its child nodes $\mathcal{C}(i)$:
\begin{equation}
\mathbf{DP}_{child, i} = \mathbf{DP}_i + \frac{1}{|\mathcal{C}(i)|} \sum_{j \in \mathcal{C}(i)} \mathbf{DP}_j
\label{eq7:dp_child}
\end{equation}
This mechanism emphasizes the local interplay between adjacent components (e.g., distinguishing the top vs. bottom part of a specific `Above-Below' structure).

\subsubsection{Multi-Source Feature Encoding}
By integrating the entropy-aware embeddings (\Cref{subsec:entropy-aware}) with the dual-view structural embeddings, we establish a comprehensive feature dictionary. Specifically, each character is represented by \textbf{five distinct encoding vectors}, which are pre-computed and stored for efficient retrieval during training and inference:

\begin{enumerate}
    \item \textbf{Entropy-Aware Representation ($\mathbf{V}_{ent}$):} The global entropy-weighted feature captures the saliency of discriminative radicals.
    \item \textbf{Radical Coding Feature ($\mathbf{F}_{code}$):} A sequence vector carrying the raw semantic information of radicals, obtained directly from the fine-tuned CLIP encoder.
    \item \textbf{Tree Depth Feature ($\mathbf{F}_{depth}$):} Encodes the hierarchical depth distribution of radicals:
    \begin{equation}
    \mathbf{F}_{depth} = \frac{\sum_{i=1}^{L} (l_i \cdot \mathbf{e}_i)}{\max(l)}
    \label{eq8:f_depth}
    \end{equation}
    \item \textbf{Global Structural Feature ($\mathbf{F}_{parent}$):} Aggregates the radical embeddings weighted by their parent-centric structural roles:
    \begin{equation}
    \mathbf{F}_{parent} = \frac{\sum_{i=1}^{L} (\mathbf{DP}_{parent, i} \odot \mathbf{e}_i)}{\|\sum_{i=1}^{L} \mathbf{DP}_{parent, i}\|_2}
    \label{eq9:f_parent}
    \end{equation}
    \item \textbf{Local Structural Feature ($\mathbf{F}_{child}$):} Aggregates the radical embeddings weighted by their child-centric local roles:
    \begin{equation}
    \mathbf{F}_{child} = \frac{\sum_{i=1}^{L} (\mathbf{DP}_{child, i} \odot \mathbf{e}_i)}{\|\sum_{i=1}^{L} \mathbf{DP}_{child, i}\|_2}
    \label{eq10:f_child}
    \end{equation}
\end{enumerate}

Since these five feature vectors rely solely on the fixed IDS and pre-trained encoders, they are extracted offline as a static dictionary. During the online inference phase, the model simply retrieves these vectors to guide the attention mechanism, significantly reducing computational overhead while maximizing structural expressiveness.

\subsection{Radical Semantic Matching Module}
Having obtained the deformation-robust visual features from the CNN backbone and the dual-view structural embeddings from the radical tree, the final and most critical challenge is to align these two heterogeneous modalities. Preliminary works typically rely on shallow metric learning, such as calculating the Cosine similarity between the global visual vector and the radical embedding. However, such linear alignment mechanisms are insufficient for zero-shot HCCR. They fail to capture the complex, non-linear correspondence between specific visual stroke patterns (e.g., a "hook" stroke) and abstract radical semantics. To address this, we propose a Multi-Stage Semantic Matching Module, which explicitly performs coarse-to-fine reasoning through adaptive feature fusion, cross-modal attention, and neighbor-aware aggregation.

\subsubsection{Adaptive GateFusion Network}\label{subsubsec:gatefusion}
\begin{figure}[t!]
    \centering
    \includegraphics[width=0.85\textwidth]{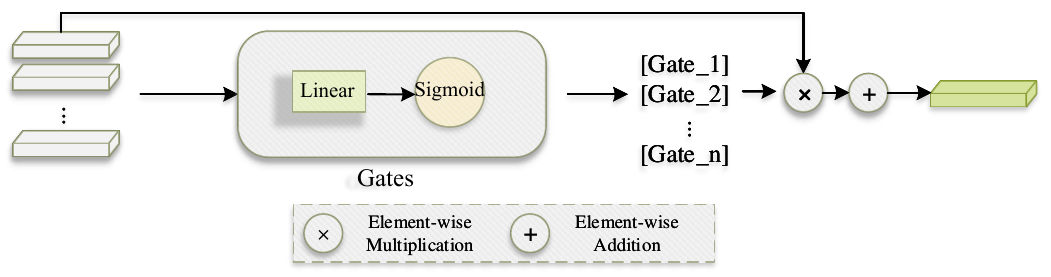}
    \vspace{-15pt}
    \caption{Architecture of the \textbf{Adaptive GateFusion Network.} The module aggregates four heterogeneous structural embeddings ($\mathbf{V}_{ent}, \mathbf{F}_{depth}, \mathbf{F}_{parent}, \mathbf{F}_{child}$) via element-wise Sigmoid gating ($\sigma$) to modulate their importance. The gated features are summed, and the radical content embedding ($\mathbf{F}_{code}$) is explicitly injected as an additive semantic bias to preserve the fundamental identity, yielding the final prototype $\mathbf{P}_{sem}$.}
    \label{fig:gatefusion}
    \vspace{-10pt}
\end{figure}

To effectively synthesize the heterogeneous structural information extracted from different views, we propose an \textbf{Adaptive GateFusion Network}. As illustrated in \Cref{fig:gatefusion}, this module functions as a comprehensive feature aggregator that dynamically balances the contribution of four key structural embeddings.

Let $\mathcal{S}$ denote the input feature set, comprising the EAPE ($\mathbf{V}_{ent}$), the Depth Embedding ($\mathbf{F}_{depth}$), and the Dual-View Radical Tree features ($\mathbf{F}_{parent}, \mathbf{F}_{child}$). Initially, to resolve the discrepancy in feature subspaces, each input feature $\mathbf{f}_i \in \mathcal{S}$ is projected into a unified high-dimensional semantic space $\mathbb{R}^d$ via a specific linear transformation $\tilde{\mathbf{f}}_i = \mathbf{W}_i \mathbf{f}_i + \mathbf{b}_i$. Subsequently, to achieve adaptive information flow, an independent gating unit employing the Sigmoid function $\sigma(\cdot)$ is applied to each aligned feature. This mechanism generates an element-wise importance map $\mathbf{g}_i = \sigma(\mathbf{W}_{gate, i} \tilde{\mathbf{f}}_i + \mathbf{b}_{gate, i})$, which modulates the feature magnitude through element-wise multiplication $\mathbf{h}_i = \mathbf{g}_i \odot \tilde{\mathbf{f}}_i$. By doing so, the network can autonomously emphasize discriminative structural cues while suppressing irrelevant noise.

In the final aggregation stage, the gated features from all four branches are synthesized via element-wise addition. Crucially, to prevent the potential dilution of the fundamental radical semantics during these structural transformations, we explicitly inject the linearly projected radical content embedding $\mathbf{F}_{code}$ as a bias term. The final comprehensive representation $\mathbf{P}_{sem}$ is thus formulated as:
\begin{equation}
    \mathbf{P}_{sem} = \sum_{\mathbf{f}_i \in \mathcal{S}} \left( \sigma(\mathbf{W}_{gate, i} \tilde{\mathbf{f}}_i + \mathbf{b}_{gate, i}) \odot \tilde{\mathbf{f}}_i \right) + \mathbf{W}_{code}\mathbf{F}_{code}
\end{equation}
This design ensures that the model effectively combines "where the radical is" (via the gated structural ensemble) with "what the radical is" (via the identity injection). We explicitly incorporate $\mathbf{F}_{depth}$ and $\mathbf{V}_{ent}$ into the gating inputs to preserve hierarchical depth signals and entropy-based importance priors, ensuring these critical physical constraints are not vanished during deep reasoning.

\subsubsection{Cross-Modal Attention Alignment}
To bridge the modality gap, we treat the alignment process as a query-retrieval task. We implement a Multi-Head Cross-Attention (MHCA) layer, where the visual features act as queries seeking relevant semantic components.

Formally, let $\mathbf{F}_{v} \in \mathbb{R}^{N_v \times d}$ denote the flattened visual feature map from the CNN backbone (where $N_v = H \times W$), and let $\mathbf{P}_{sem} \in \mathbb{R}^{L \times d}$ denote the sequence of fused semantic prototypes for a candidate character class (where $L$ is the radical sequence length). We define the Query ($\mathbf{Q}_{sem}$), Key ($\mathbf{K}$), and Value ($\mathbf{V}$) matrices as:
\begin{equation}
\mathbf{Q}_{sem} = \mathbf{P}_{sem} \mathbf{W}_q, \quad \mathbf{K} = \mathbf{F}_{v} \mathbf{W}_k, \quad \mathbf{V} =  \mathbf{F}_{v}\mathbf{W}_v
\label{eq13:qkv}
\end{equation}
The cross-attention map is computed to measure the affinity between each visual region and each radical prototype:
\begin{equation}
\text{Attention}_{sem}(\mathbf{Q}_{sem}, \mathbf{K}, \mathbf{V}) = \text{softmax} \left( \frac{\mathbf{Q}_{sem}\mathbf{K}^T}{\sqrt{d_k}} \right) \mathbf{V}
\label{eq14:atten_qkv}
\end{equation}
By performing this cross-attention, the model generates a visually-grounded semantic representation $\hat{\mathbf{S}}$. This explicitly forces the network to "look at" the specific image regions (e.g., the left part of the image) that correspond to the target radical (e.g., the left radical in the IDS), significantly suppressing background noise and stroke artifacts.

Finally, the matching score $Score(x, y_c)$ is computed by measuring the similarity (e.g., Euclidean distance or Cosine similarity) between the global visual token and the aggregated output of the cross-attention module.

\subsubsection{Top-K Semantic Feature Fusion}
\begin{CJK*}{UTF8}{gbsn}
In ZSL scenarios, relying solely on the single most relevant semantic vector (Top-1) retrieved by cross-modal alignment is often brittle. Due to the high-dimensional nature of the shared embedding space and the subtle structural differences between Chinese characters (e.g., ``盲'' (blind) vs. ``育'' (educate)), the mathematically nearest neighbor may not always correspond to the semantic ground truth.
\end{CJK*}

To enhance the fault tolerance of the model, we propose a Top-K Semantic Feature Fusion strategy. Unlike post-hoc score averaging, this module operates at the feature level, leveraging the semantic consensus of the top-$K$ nearest radical prototypes to construct a robust query for the Transformer decoder.

Specifically, after computing the cross-modal affinity scores between the visual feature and all candidate semantic prototypes, we identify the indices of the top-$K$ candidates, denoted as $\mathcal{N}_K$. We then retrieve their corresponding fused semantic vectors $\{\mathbf{p}_{sem}^{(k)} \mid k \in \mathcal{N}_K\}$. To mitigate the noise from potential outliers in the retrieval set, we perform a mean fusion operation to generate a robust semantic prototype $\mathbf{P}_{robust}$:
\begin{equation}
\mathbf{P}_{robust} = \frac{1}{K} \sum_{k \in \mathcal{N}_K} \mathbf{Attention}_{sem}^{(k)}
\label{eq14:p_robust}
\end{equation}
This aggregated prototype represents the centroid of the semantic neighborhood. Finally, to guide the decoding process with this enhanced prior, we inject $\mathbf{P}_{robust}$ into the visual pipeline by adding it to the decoder's query vector $\mathbf{Q}$:
\begin{equation}
\mathbf{Q}' = \mathbf{Q} + \mathbf{P}_{robust}
\label{eq15:q}
\end{equation}
By integrating the intersection of radical information from these $K$ neighbors (which likely share common components with the target), the model can hallucinate the correct structural identity even if the Top-1 match is partially incorrect. Empirical results demonstrate that $K=5$ yields the optimal performance, aligning with the statistical observation that the average radical sequence length of Chinese characters is approximately 5.

\section{Experiments}
In this section, we conduct a comprehensive evaluation of the proposed Entropy-Aware Structural Alignment Network. We first detail the datasets and experimental protocols, followed by a comparative analysis against state-of-the-art (SOTA) methods on both zero-shot and few-shot recognition tasks. Furthermore, we perform extensive ablation studies to systematically isolate and verify the contributions of key components. Finally, qualitative visualizations are provided to offer deeper insights into the model's internal reasoning mechanism.

\subsection{Experimental Setup}
\subsubsection{Datasets and Protocols}
We evaluate our method on the standard \textbf{CASIA-HWDB} (1.0/1.1) \cite{liu2011} and \textbf{ICDAR 2013} \cite{yin2013} benchmarks. Following the strict zero-shot protocol established in prior works \cite{huang2022b,pan2025}, we focus on the 3,755 Level-1 GB2312 characters, partitioning them into two disjoint sets ($\mathcal{Y}_{seen} \cap \mathcal{Y}_{unseen} = \emptyset$). The \textbf{Source Set (Seen)} comprises 2,755 classes from CASIA-HWDB for training, ensuring full coverage of atomic radicals. The \textbf{Target Set (Unseen)} reserves the remaining 1,000 classes for zero-shot evaluation using samples from ICDAR 2013. We report Top-1 Character Recognition Accuracy (Acc@1) to quantify performance.

\subsubsection{Implementation Details}
Our framework is implemented in PyTorch on four NVIDIA RTX 4090 GPUs. The visual stream employs a randomly initialized ResNet-34 backbone, which is trained from scratch on the training set to learn domain-specific handwritten features. The semantic radical encoder is fine-tuned using weights from the CLIP (ViT-B/32) image encoder. This fine-tuning process is conducted prior to the main backbone training, following the protocol detailed in our preliminary work \cite{luo2025}. Consequently, all radical features are extracted via this pre-trained CLIP encoder to construct a static feature dictionary. During the main network training, the CLIP parameters are frozen, serving solely for efficient vector retrieval. The entire model is optimized end-to-end using the Adadelta optimizer with a batch size of 32. We employ a MultiStepLR scheduler, where the learning rate is initialized to 0.1 and decayed to 0.01 and 0.001 at the 15th and 20th epochs, respectively. Regarding key hyperparameters, the neighbor count $K$ for the Top-K Semantic Feature Fusion strategy is set to 5 based on our ablation studies.

\subsection{Comparison with State-of-the-Art}
In this subsection, we conduct a comprehensive evaluation of the proposed framework against SOTA methods. To demonstrate the versatility and robustness of our approach, the comparison is performed across three distinct dimensions: (1) zero-shot generalization on unseen characters, (2) fully supervised recognition on the complete dataset, and (3) inference efficiency in practical deployment.

\subsubsection{Performance on Zero-shot Character Recognition}
The core contribution of this work is the ability to recognize unseen characters by leveraging radical-based semantic priors. We evaluate our method against a comprehensive list of SOTA approaches. \Cref{tab:sota_comparison} summarizes the Top-1 accuracy on the ICDAR 2013 target set (Unseen) under the zero-shot setting, where the number of seen training classes $m$ varies from 500 to 2,755. Our Entropy-Aware Structural Alignment Network establishes a new SOTA, consistently outperforming all existing methods across different data regimes.

\begin{table}[t]
    \centering
    \caption{Comparison with State-of-the-Art Methods on zero-shot Chinese Character and Radical Recognition. The metric is Top-1 Recognition Accuracy (\%).}
    \label{tab:sota_comparison}
    \fontsize{9pt}{8pt}\selectfont
    \setlength{\tabcolsep}{2.5pt}
    \begin{tabular}{llcccccccccc} 
        \toprule
        \multirow{2}{*}{\textbf{Method}} & \multirow{2}{*}{\textbf{Year}} & \multicolumn{5}{c}{\textbf{Number of Unseen characters} ($m$)} & \multicolumn{5}{c}{\textbf{Number of Unseen Radical} ($n$)} \\
        \cmidrule(lr){3-7} \cmidrule(lr){8-12}
         & & \textbf{500} & \textbf{1000} & \textbf{1500} & \textbf{2000} & \textbf{2755} & \textbf{50} & \textbf{40} & \textbf{30} & \textbf{20} & \textbf{10} \\
        \midrule
        HDE\cite{cao2020} & 2020 & 4.90 & 12.77 & 19.25 & 25.13 & 33.49 & 3.26 & 4.29 & 6.33 & 7.64 & 9.33 \\
        SLD\cite{chen2021a} & 2021 & 5.60 & 13.85 & 22.88 & 25.73 & 37.91 & 5.28 & 6.87 & 9.02 & 14.76 & 15.83 \\
        ACPM\cite{zu2022} & 2022 & 9.72 & 18.50 & 27.74 & 34.00 & 42.43 & - & - & - & - & - \\
        STAR\cite{zeng2023b} & 2023 & 7.54 & 19.47 & 27.79 & 35.53 & 43.86 & 6.95 & 12.28 & 14.74 & 18.37 & 23.23 \\
        SIR\cite{luo2023b} & 2023 & 7.43 & 15.75 & 24.01 & 27.04 & 40.55 & - & - & - & - & - \\      
        CCR-CLIP\cite{yu2023a} & 2023 & 21.03 & 38.43 & 48.85 & 52.70 & 62.59 & 21.16 & 32.16 & 38.86 & 46.00 & 57.21 \\
        SideNet\cite{li2024o} & 2024 & 5.10 & 16.25 & 33.80 & 44.10 & 50.30 & - & - & - & - & - \\
        RSST\cite{yu2024e} & 2024 & 11.56 & 21.83 & 35.32 & 39.22 & 47.44 & 7.94 & 11.56 & 15.13 & 15.92 & 20.21 \\
        HierCode\cite{zhang2025k} & 2025 & 6.22 & 20.71 & 35.39 & 45.67 & 56.21 & - & - & - & - & - \\
        UCR-RRM\cite{li2025e} & 2025 & 15.16 & 24.04 & 37.46 & 42.21 & 49.36 & 10.57 & 14.34 & 17.68 & 20.05 & 23.41 \\
        RSE-CLIP\cite{luo2025} & 2025 & 9.03 & 17.71 & 35.02 & 48.96 & 59.80 & 19.15 & 21.09 & 28.45 & 35.80 & 43.42 \\
        \midrule
        \textbf{Ours} & 2026 & \textbf{24.54} & \textbf{43.13} & \textbf{55.04} & \textbf{60.50} & \textbf{67.96} & \textbf{24.61} & \textbf{34.93} & \textbf{44.11} & \textbf{52.48} & \textbf{60.20} \\
        \bottomrule
    \end{tabular}
    \vspace{-10pt}
\end{table}

\textbf{Superiority over Structural and Decomposition Models:}
Our method demonstrates a substantial lead over traditional decomposition approaches (e.g., +34.47\% over HDE and +25.53\% over ACPM), proving that static radical mapping is insufficient for handwritten variance. Furthermore, compared to the latest structural models from 2024 to 2025, our approach significantly outperforms the stroke-based RSST by +20.52\% and the hierarchical HierCode by +11.75\%. These results highlight a critical insight: while strokes and standard trees capture topology, they often introduce noise. Our Information Entropy Prior effectively suppresses such ubiquitous structural noise, allowing the model to focus on discriminative semantic roots, which proves more robust than stroke-level or unweighted hierarchical features.

\textbf{Advancement over CLIP-Enhanced and Preliminary Works:}
A critical comparison is drawn against CLIP-based baselines to validate our specific contributions. Our method surpasses the strong CCR-CLIP by +5.37\%, confirming that explicitly modeling the parent-child topology is superior to flat visual-semantic alignment. Most importantly, compared to our preliminary conference version RSE-CLIP (59.80\%), which shares the same structural backbone but lacks the entropy mechanism, our proposed framework achieves a remarkable +8.16\% improvement. This significant performance leap serves as a direct validation, confirming that the introduction of Entropy-Aware Saliency and Adaptive Top-K Fusion is the decisive factor in resolving the ambiguity of unseen characters.

\subsubsection{Generalization to Full-Set Recognition}
While our primary focus is zero-shot learning, we also evaluate the proposed framework on the standard fully supervised setting to verify its general applicability. \Cref{tab:full_set_recognition} compares our method with SOTA fully supervised approaches based on their auxiliary supervision (e.g., Glyphs, Radicals) and similarity metrics. As shown, our method achieves a highly competitive accuracy of \textbf{97.30\%}, demonstrating that the proposed entropy-aware structural alignment remains robust and effective even when abundant training data is available.

\begin{table}
    \centering
    \caption{Comparison of Full-Set Recognition Accuracy. Methods are compared based on the types of auxiliary information utilized and the similarity metric strategy employed.}
    \label{tab:full_set_recognition}
    \fontsize{9pt}{8pt}\selectfont
    \setlength{\tabcolsep}{3pt}
    \begin{tabular}{l l l l c}
        \toprule
        \textbf{Method} & \textbf{Year} & \textbf{Auxiliary Info} & \textbf{Metric Strategy} & \textbf{Acc. (\%)} \\ 
        \midrule
        DenseRAN\cite{wang2018} & 2018 & Radical & N/A & 96.66 \\
        FewshotRAN\cite{wang2019} & 2019 & Radical & N/A & 96.97 \\
        TemplateLoss\cite{xiao2019} & 2019 & Glyph & Variance Constraint & 97.45 \\
        HDE\cite{cao2020} & 2020 & Radical & Cosine Similarity & 97.14 \\
        SLD\cite{chen2021a} & 2021 & Radical+Stroke & Cosine Similarity & 96.28 \\
        ACPM\cite{zu2022} & 2022 & Glyph+Radical+Stroke & Fixed Distance & \textbf{97.80} \\
        CCR-CLIP\cite{yu2023a} & 2023 & Glyph+Radical & Cosine Similarity & 97.18 \\    
        TPE\cite{xue2023a} & 2023 & Radical & Cosine + MSE & 96.88 \\
        LERRNet\cite{pan2025} & 2025 & Radical & Cross-Modal Attention & 97.39 \\
        \midrule
        \textbf{Ours} & \textbf{2026} & \textbf{Radical} & \textbf{Cross-Attn + Top-K Agg.} & \textbf{97.30} \\ 
        \bottomrule
    \end{tabular}
    \vspace{-10pt}
\end{table}

\textbf{Comparison with Multi-Modal Heavyweights:}
The current top-performing method, ACPM (97.80\%), achieves a marginal lead of 0.5\% but relies on a complex fusion of three modalities: Glyphs (images), Radicals, and Strokes. This reliance on glyph templates significantly increases the training data requirement and computational burden. In contrast, our method utilizes only Radical information yet delivers comparable performance. This demonstrates the superior data efficiency of our Entropy-Aware Structural modeling—we achieve near-SOTA accuracy without the overhead of processing pixel-level glyph templates or detailed stroke sequences.

\textbf{Comparison with Radical-based Peers:} Among methods that similarly rely on radical information (e.g., DenseRAN, HDE, TPE), our method outperforms most competitors. Notably, we surpass our baseline HDE (97.14\%) and CCR-CLIP (97.18\%), proving that our structural alignment is more effective than simple cosine similarity.

The results also highlight the importance of the matching mechanism. Most existing methods (e.g., HDE, SLD) employ Cosine Similarity, which is computationally efficient but often fails to capture complex non-linear cross-modal dependencies. By introducing Cross-Modal Attention followed by Top-K Aggregation, our framework can better align visual features with semantic embeddings, leading to more robust recognition than fixed-distance metrics.

\subsubsection{Inference Efficiency Analysis}
Beyond recognition accuracy, computational efficiency is critical for real-world deployment. To ensure a fair comparison with prior works such as UCR \cite{li2025e}, we benchmark the inference speed on a single NVIDIA RTX 4090 GPU. Following standard protocols, the input batch size is set to 1, and the image resolution is fixed at $32 \times 32$ pixels.

\begin{table}
    \centering
    \caption{Comparison of Inference Efficiency.}
    \label{tab:inference_time}
    \fontsize{10pt}{8pt}\selectfont
    \begin{tabular}{llcc}
        \toprule
        \textbf{Method} & \textbf{Year} & \textbf{Input Type} & \textbf{Speed (ms)} \\ 
        \midrule
        DenseNet \cite{huang2017} & 2017 & Character & 0.76 \\
        RAN \cite{zhang2020} & 2020 & Radical & 0.62 \\
        ViT \cite{dosovitskiy2021} & 2021 & Character & 1.29 \\
        RTN \cite{yang2021b} & 2021 & Radical & 0.92 \\
        SLD \cite{chen2021a} & 2021 & Stroke & 1.21 \\
        HCRN \cite{huang2022b} & 2022 & Radical & 1.00 \\
        UCR \cite{li2025e} & 2025 & Wubi+Radical & 1.43 \\  
        \midrule 
        \textbf{Ours} & \textbf{2026} & \textbf{Radical} & \textbf{0.74} \\ 
        \bottomrule
    \end{tabular}
    \vspace{-10pt}
\end{table}

As shown in \Cref{tab:inference_time}, our method achieves a highly efficient inference speed of 0.74 ms per image. \textbf{Comparable to Simple CNNs:} Remarkably, our speed is on par with the classical character-based classification model DenseNet (0.76 ms). This indicates that incorporating fine-grained radical structural reasoning does not impose a significant latency overhead compared to standard black-box classifiers. \textbf{Superior to Complex Multi-Modal Models:} Compared to the latest framework UCR (2025, 1.43 ms), our method is nearly 2$\times$ faster. While UCR integrates auxiliary Wubi encodings to boost performance, this multi-modal fusion significantly increases computational complexity. In contrast, our method relies solely on radicals and achieves superior efficiency. Similarly, compared to stroke-based models like SLD (1.21 ms), our radical-level modeling avoids the lengthy sequence decoding of strokes. \textbf{Optimal Speed-Accuracy Trade-off:} Although RAN (0.62 ms) is slightly faster, it employs a relatively simple attention mechanism that limits its capability to capture complex structural variations, resulting in lower recognition accuracy. Our method strikes an optimal balance: it maintains the high speed of lightweight models while delivering SOTA zero-shot accuracy through our efficient Offline-Online Decoupling strategy. The heavy computations (Tree Parsing, Entropy Calculation) are pre-loaded as static lookups, leaving only the visual backbone and lightweight retrieval for online inference.

\subsection{Ablation Studies}
To systematically verify the contribution of each proposed component, we conduct extensive ablation studies on the ICDAR 2013 zero-shot dataset. We decompose the evaluation of the Entropy-Aware Structural Alignment Network into three critical aspects: (1) Radical Position Embedding strategies, (2) Semantic Matching mechanisms, and (3) Hyperparameter Sensitivity regarding the fusion scope $K$.

\subsubsection{Effectiveness of EAPE}
To isolate the contribution of the proposed EAPE, we conduct a progressive ablation study comparing it against standard positional encoding schemes on the zero-shot radical recognition task. The baseline employs a simple one-hot radical encoding without any structural position information. We then incrementally incorporate the radical tree topology and various position embedding strategies to measure their specific impacts.

The quantitative results are summarized in \Cref{tab:ablation_pe}. As observed, the introduction of the Radical Tree topology alone (without explicit position indices) triggers the most significant performance leap, boosting the accuracy from 16.53\% to 23.23\% at $m=500$. This confirms that the hierarchical connection itself provides essential structural context that a flat baseline lacks. However, simply adding standard geometric encodings—such as Sinusoidal PE or RoPE—to this tree structure yields only marginal gains (e.g., Sinusoidal PE improves by merely $\sim$0.4\% over the tree-only baseline). This saturation suggests that treating character structure purely as a geometric sequence is suboptimal, as it fails to account for the varying semantic importance of different nodes. In contrast, our Entropy-Aware mechanism effectively breaks this bottleneck, achieving the highest accuracy of 24.54\% ($m=500$) and 55.04\% ($m=1500$). By utilizing multiplicative interaction to modulate the positional embeddings, our method amplifies the feature magnitude of high-entropy radicals while dampening ubiquitous particles.

\begin{table}
    \centering
    \caption{Ablation Study on Position Embedding Strategies.}
    \label{tab:ablation_pe}
    \fontsize{10pt}{8pt}\selectfont
    \begin{tabular}{lcc}
        \toprule
        \multirow{2}{*}{\textbf{Position Embedding Strategy}} & \multicolumn{2}{c}{\textbf{Accuracy (\%)}} \\
        \cline{2-3}
         & \textbf{m = 500} & \textbf{m = 1500} \\
        \midrule
        Baseline (One-hot w/o Tree) & 16.53 & 48.85 \\
        + Radical Tree (Structure Only) & 23.23 & 53.72 \\
        + Sinusoidal PE (Standard) & 23.61 & 54.25 \\
        + RoPE (Rotary Embedding) & 23.95 & 54.50 \\
        \midrule
        \textbf{+ Entropy-Aware PE} & \textbf{24.54} & \textbf{55.04} \\
        \bottomrule
    \end{tabular}
    \vspace{-10pt}
\end{table}

\subsubsection{Efficacy of Semantic Matching Strategies}
We investigate the impact of the matching mechanism, which aligns visual features with the learned structural embeddings. To validate the superiority of our Multi-Stage Semantic Matching module (Cross-Modal Attention + Top-K Fusion), we compare it against deterministic metrics ($L1/L2$ distance) and standard similarity measures (Cosine Similarity) widely used in CLIP-based methods.

The quantitative results in \Cref{tab:ablation_matching} reveal the limitations of traditional metrics in zero-shot scenarios. Simple geometric distances ($L1$ and $L2$) yield the lowest performance (approx. 21.4\% at $m=500$), indicating that the high-dimensional visual-semantic manifold is too complex for linear separation. While standard Cosine Similarity improves the accuracy to 21.83\%, it still lags behind our proposed method. In contrast, integrating the Cross-Modal Attention layer significantly boosts performance to 24.54\% (+2.71\%). This substantial gain confirms that the adaptive attention mechanism can better capture the non-linear dependencies between distorted handwritten strokes and canonical radical definitions. Furthermore, the ablation variant "Softmax + Top-K" (removing the attention layer) results in a performance drop to 21.09\%, demonstrating that the Top-K feature fusion strategy relies heavily on the refined feature alignment provided by the attention module to function effectively.

\begin{table}[t]
    \centering
    \caption{Ablation Study on Semantic Matching Strategies. Comparison of zero-shot recognition accuracy (\%) under different metric strategies.}
    \label{tab:ablation_matching}
    \fontsize{10pt}{8pt}\selectfont
    \begin{tabular}{lcc}
        \toprule
        \multirow{2}{*}{\textbf{Matching Strategy}} & \multicolumn{2}{c}{\textbf{Accuracy (\%)}} \\
        \cline{2-3}
         & \textbf{m = 500} & \textbf{m = 1500} \\
        \midrule
        L1 Distance & 21.41 & 51.88 \\
        L2 Distance & 21.46 & 51.51 \\
        Cosine Similarity & 21.83 & 54.13 \\
        Softmax + Top-K & 21.09 & 54.15 \\
        \midrule
        \textbf{Cross-Attention + Top-K} & \textbf{24.54} & \textbf{55.04} \\
        \bottomrule
    \end{tabular}
    \vspace{-10pt}
\end{table}

\subsubsection{Hyperparameter Sensitivity: Analysis of $K$}\label{subsubsec:param_k}
The hyperparameter $K$ in our Top-K Semantic Feature Fusion module governs the scope of semantic neighborhood reasoning. To determine the optimal value, we analyze the zero-shot accuracy ($m=500$) by varying $K$ from 1 to 7.

\begin{figure}[t!]
    \centering
    \includegraphics[width=0.7\textwidth]{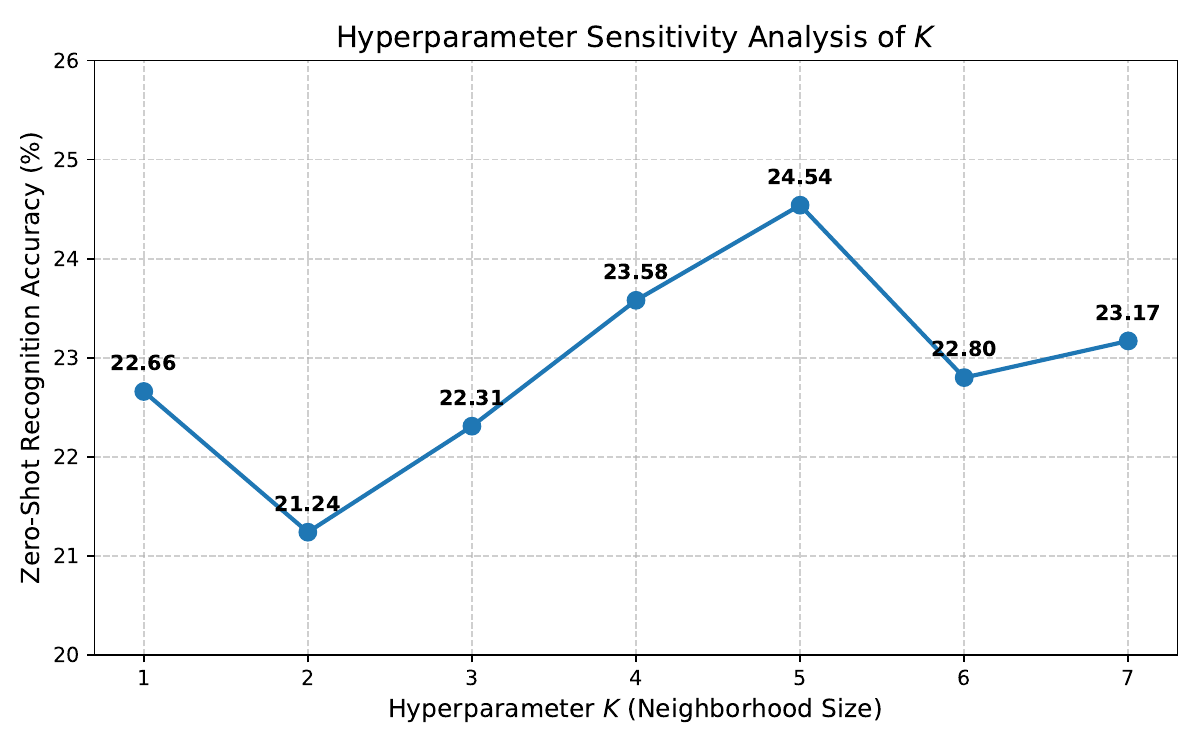} 
    \vspace{-10pt}
    \caption{Sensitivity analysis of hyperparameter $K$.}
    \label{fig:ablation_k}
    \vspace{-10pt}
\end{figure}

As visually illustrated in \Cref{fig:ablation_k}, the recognition accuracy does not follow a simple monotonic trend. The 1-Nearest Neighbor baseline ($K=1$) starts with a relatively high accuracy of 22.66\%, but increasing $K$ to 2 causes a counter-intuitive drop to 21.24\%. This phenomenon suggests that in a sparse zero-shot embedding space, the second-closest candidate is often a distractor rather than a valid semantic variant, thereby introducing noise into the fused feature representation. However, as $K$ increases further from 3 to 5, the performance rebounds and climbs steadily, peaking at 24.54\% when $K=5$. This indicates that expanding the fusion pool allows the semantic centroid of true neighbors to outweigh outlier noise, demonstrating the robustness of our feature fusion strategy. Beyond this peak, when $K$ exceeds 5, the accuracy begins to degrade, confirming that an overly broad scope incorporates irrelevant classes that confuse the decision boundary. Notably, the optimal value $K=5$ coincides with the average radical sequence length of Chinese characters (approx. 5.01) in our dataset. This statistical alignment provides a compelling physical interpretation: the model achieves optimal robustness when the feature fusion scope matches the intrinsic structural complexity of the characters.

\subsection{Analysis of Few-Shot Adaptability and Data Efficiency}
\begin{figure}[t!]
    \centering
    \includegraphics[width=0.7\textwidth]{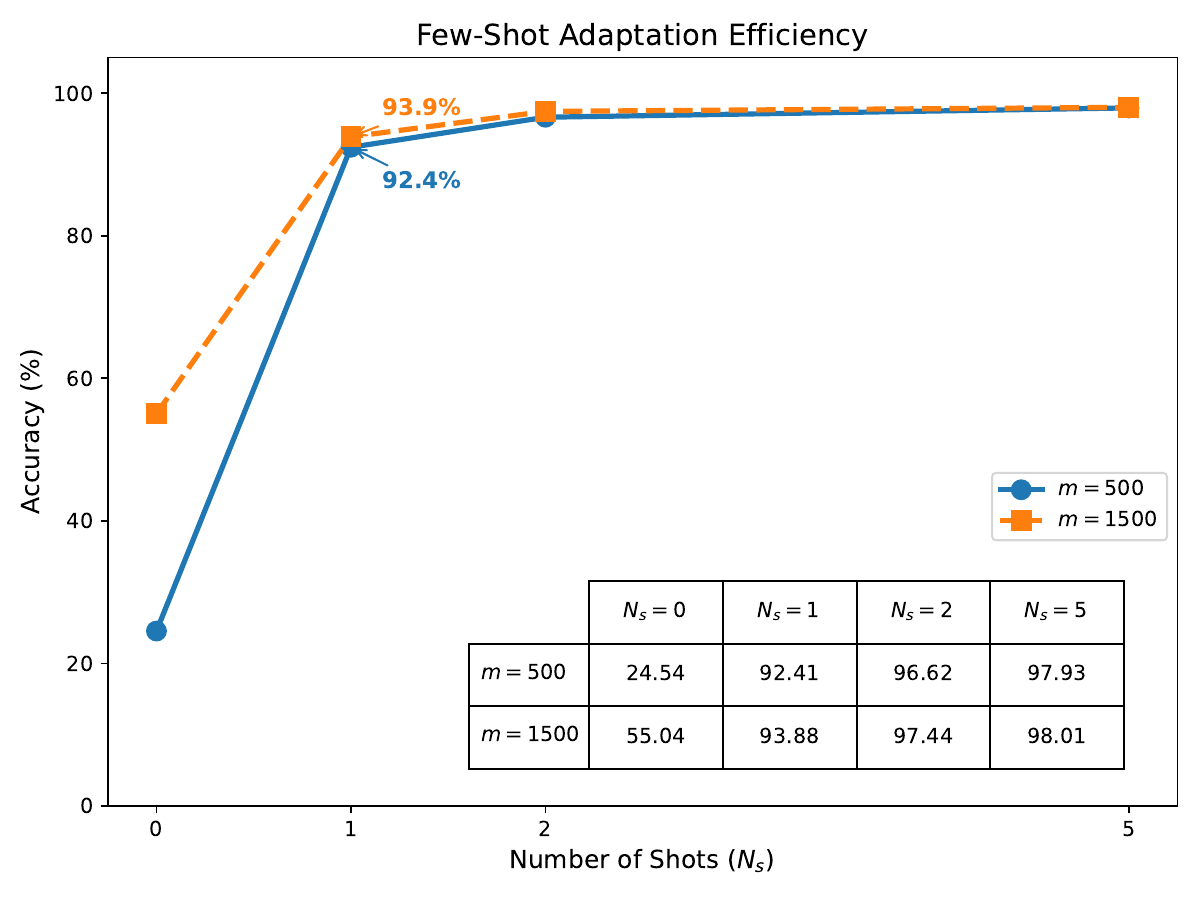} 
    \vspace{-10pt}
    \caption{Evaluation of Few-Shot Adaptability.}
    \label{fig:few_shot_trend}
    \vspace{-10pt}
\end{figure}

While zero-shot recognition demonstrates the model's ability to reason from semantic definitions, real-world applications often permit a "cold start" with a minimal number of support samples. To evaluate the data efficiency of our framework, we investigate the performance gains when providing $N_s$ reference samples per unseen category, specifically setting $N_s \in \{1, 2, 5\}$. The performance trends and detailed numerical results are presented in \Cref{fig:few_shot_trend}.

\textbf{Rapid Adaptation Performance:} Notably, a single visual prototype ($N_{s}=1$) triggers a massive leap: for the $m=500$ split, accuracy surges from 24.54\% (Zero-shot) to \textbf{92.41\%}, effectively bridging the majority of the domain gap. Increasing to $N_{s}=2$ further refines accuracy to \textbf{96.62\%}, while $N_{s}=5$ shows clear saturation at \textbf{97.93\%}. This steep ascent followed by diminishing returns confirms our model's high sample efficiency, achieving robust recognition with as few as two examples. \textbf{Mechanism of Efficiency:} We attribute this capability to the \textbf{Entropy-Aware Structural Prior}. Unlike conventional few-shot learning, which requires multiple samples to statistically infer discriminative features ("Statistical Exploration"), our framework operates with a pre-computed "structural blueprint." The information entropy explicitly highlights high-value radicals beforehand, allowing the attention mechanism to immediately "lock onto" critical regions upon seeing the first sample ($N_{s}=1$). This effectively transforms the learning process into direct "Structural Registration," enabling precise alignment with minimal visual evidence.

\subsection{Visualization of Cross-Modal Reasoning Process}
\begin{figure}[t!]
    \centering
    \includegraphics[width=0.9\textwidth]{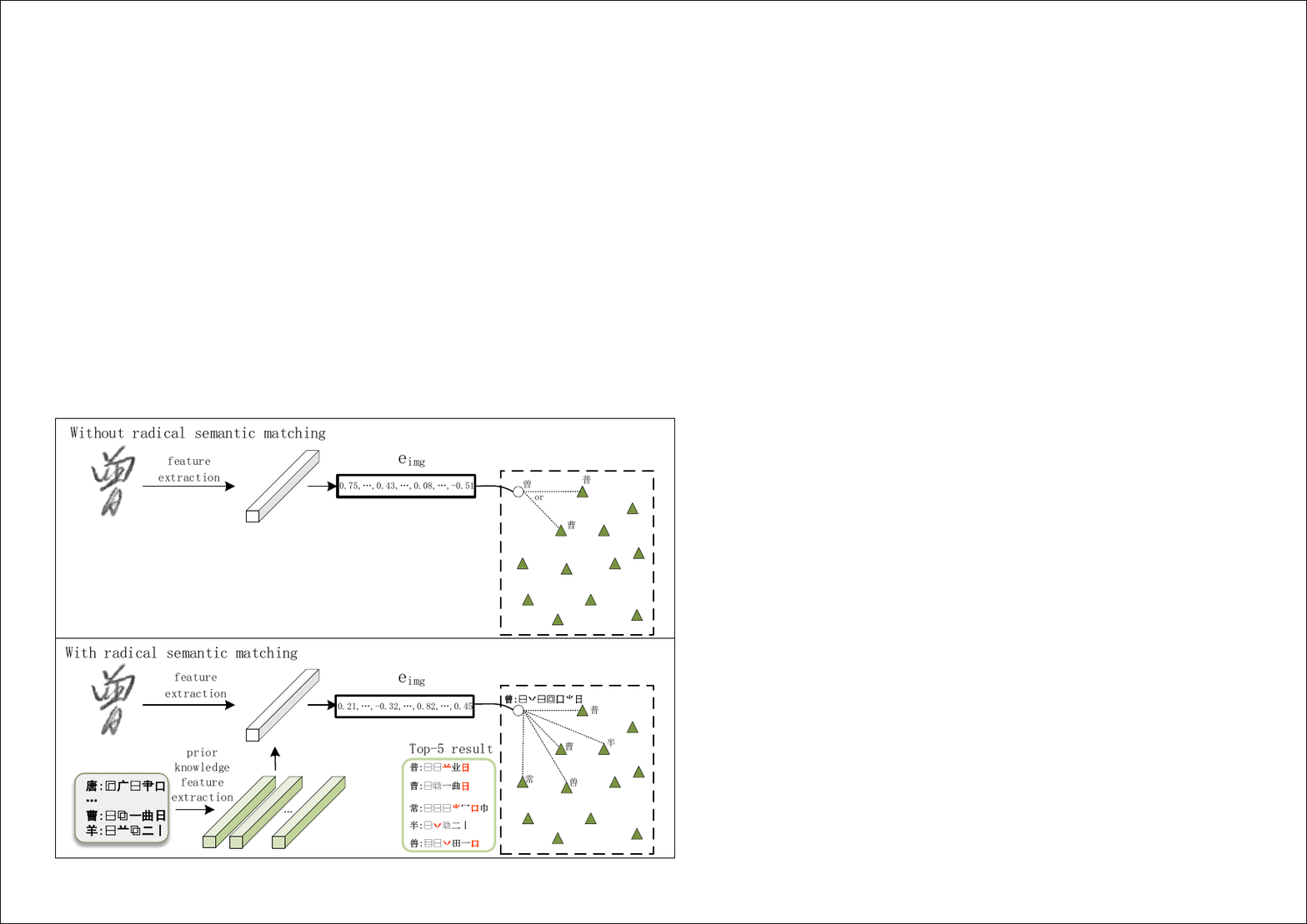}
    \vspace{-10pt}
    \begin{CJK*}{UTF8}{gbsn}
    \caption{Visualization of the reasoning process for the unseen character ``曾''. The baseline fails due to visual ambiguity with shape-similar characters. Our method leverages radical priors. Notably, while a single neighbor may only match partial radicals, the Top-5 neighbors collectively provide the complete set of necessary radicals (highlighted in red), allowing the Soft Aggregation module to correctly reconstruct the target identity.}
    \label{fig:vis_reasoning}
    \end{CJK*}
    \vspace{-10pt}
\end{figure}

\begin{CJK*}{UTF8}{gbsn}
To demystify the cross-modal reasoning, we visualize the inference of a challenging unseen character ``曾'' in \Cref{fig:vis_reasoning}. The baseline, relying solely on global visual features, fails to capture internal topology and misclassifies it as morphologically similar characters (e.g., ``普'' or ``曹''). In contrast, our Entropy-Aware network imposes radical-level constraints. Crucially, the visualization reveals that while a single Top-1 retrieval may only match partial components, the Top-5 semantic neighbors (including ``普'', ``曹'', and ``半'') collectively contain all the constituent radicals required to construct ``曾'' (highlighted in red). The Soft Aggregation mechanism effectively assembles these dispersed semantic fragments, successfully reconstructing the full target identity from the consensus of local structural cues.
\end{CJK*}

\section{Conclusion}
In this paper, we presented the \textbf{Entropy-Aware Structural Alignment Network}, a novel framework designed to bridge the gap between seen and unseen classes in zero-shot and few-shot Handwritten Chinese Character Recognition. Addressing the limitations of flat sequence modeling and rigid template matching, we introduced three key innovations: an information-theoretic entropy modulation strategy, a dual-view multi-granularity structural encoding approach, and a Top-K semantic feature fusion mechanism. Our extensive experiments yield several critical insights: (1) \textbf{Entropy as a Multiplicative Modulator:} We demonstrated that the informational value of radicals follows a heavy-tailed distribution. By modulating positional embeddings via multiplicative interaction based on information entropy, the model successfully prioritizes discriminative roots over ubiquitous particles. This approach proves superior to simple additive embeddings in suppressing high-frequency redundant signals. (2) \textbf{Adaptive Structural Fusion:} The ablation studies confirm that extracting five distinct multi-granularity features (from radical codes to global/local tree views) provides a comprehensive structural description. Furthermore, the adaptive Sigmoid-based GateFusion mechanism effectively balances these heterogeneous cues, allowing the model to dynamically shift focus between global topology and local details based on character complexity. (3) \textbf{Robustness via Feature Fusion:} The visualization of the reasoning process reveals that our Top-K Semantic Feature Fusion strategy effectively mitigates the brittleness of Top-1 retrieval. By injecting the centroid of the semantic neighborhood into the decoder's query vector, the model achieves a "structural error correction", reconstructing the correct identity even when the immediate visual-semantic alignment is ambiguous.

\textbf{Future Work:}
While our framework proves effective for isolated character recognition, applying it to end-to-end text line recognition remains a challenging frontier, primarily due to the complexity of character segmentation in cursive streams. Additionally, exploring the applicability of the Entropy-Aware mechanism to other ideographic languages  or fine-grained object recognition tasks represent a promising direction for future research.

\section{CRediT authorship contribution statement}
\textbf{Qiuming Luo:} Conceptualization, Methodology, Resources, Supervision. 
\textbf{Tao Zeng:} Investigation, Software, Formal analysis. 
\textbf{Feng Li:} Data Curation, Funding acquisition. 
\textbf{Heming Liu:} Visualization, Funding acquisition. 
\textbf{Rui Mao:} Validation, Writing - Review \& Editing. 
\textbf{Chang Kong:} Writing - Original Draft, Project administration, Supervision.

\section{Data availability}
This work used  publicly datasets.

\section{Declaration of competing interest}
The authors declare that they have no known competing financial interests or personal relationships that could have appeared to influence the work reported in this paper. 

\section{Acknowledgements}
This work was supported by the Shenzhen Key Laboratory of Embedded System Design, the Shenzhen Key Laboratory of Service Computing and Applications, the Post-doctoral Later-stage Foundation Project of Shenzhen Polytechnic University (Grant No. 6023271039K), and the Key Scientific Research Project of Shenzhen Polytechnic University (Grant No. 6024310004K).

\bibliographystyle{elsarticle-num} 
\bibliography{refs}

\end{document}